\documentclass[letterpaper, 10 pt, conference]{ieeeconf}  

\IEEEoverridecommandlockouts                              

\usepackage{graphics} 

\usepackage{amsmath}
\usepackage{amssymb}
\usepackage{multicol}
\usepackage{graphicx}
\usepackage{layouts}
\usepackage{wrapfig}
\usepackage{caption}
\usepackage{graphicx,subcaption}
\usepackage{easyReview}
\usepackage{printlen}
\usepackage{bm}
\usepackage{siunitx}
\usepackage{colortbl}	
\usepackage[utf8]{inputenc}
\usepackage{pgf}
\usepackage{tikz}
\usepackage{tikzscale}
\usepackage{pgfplots}
\usepackage[hidelinks]{hyperref}
\usepackage{placeins}
\usepackage{dblfloatfix}
\usepackage{pgf}

\pgfplotsset{compat=1.14}
\usepgfplotslibrary{external}
\usepgfplotslibrary[external]
\usepackage{multirow}
\tikzset{external/force remake}

\newcommand{\gencoor}{\boldsymbol{q}}
\newcommand{\genvel}{\boldsymbol{v}}
\newcommand{\nonlinear}{\boldsymbol{n}}
\newcommand{\torque}{\boldsymbol{\tau}}
\newcommand{\force}{\boldsymbol{\lambda}}
\newcommand{\nullspace}{P}
\newcommand{\inertial}{\boldsymbol{\phi}}
\newcommand{\identity}{\mathbf{1}}
\newcommand{\com}{\boldsymbol{c}}

\title{\LARGE \bf
Physically-Consistent Parameter Identification of Robots in Contact}

\author{
Shahram Khorshidi$^{1}$, Murad Dawood$^{1}$, Benno Nederkorn$^{2,3}$, Maren Bennewitz$^{1}$, Majid Khadiv$^{3}$
\thanks{$^{1}$Authors are with the Humanoid Robots Lab, University of Bonn, Germany, and additionally with the Lamarr Institute for Machine Learning and Artificial Intelligence and the Center for Robotics, Bonn, Germany.}
\thanks{$^{2}$Roboverse Reply, Munich, Germany.}
\thanks{$^{3}$Munich Institute of Robotics and Machine Intelligence (MIRMI), Technical University of Munich (TUM), Germany. {\tt\small majid.khadiv@tum.de}}
\thanks{This work has been partially funded by the EC, grant No. 964854, and by the BMBF within the Robotics Institute Germany, grant No. 16ME0999.}%
}

\begin{document}

\maketitle
\thispagestyle{empty}
\pagestyle{empty}

\begin{abstract}
Accurate inertial parameter identification is crucial for the simulation and control of robots encountering intermittent contacts with the environment. Classically, robots' inertial parameters are obtained from CAD models that are not precise (and sometimes not available, e.g., Spot from Boston Dynamics), hence requiring identification. To do that, existing methods require access to contact force measurement, a modality not present in modern quadruped and humanoid robots. This paper presents an alternative technique that utilizes joint current/torque measurements —a standard sensing modality in modern robots— to identify inertial parameters without requiring direct contact force measurements. By projecting the whole-body dynamics into the null space of contact constraints, we eliminate the dependency on contact forces and reformulate the identification problem as a linear matrix inequality that can handle physical and geometrical constraints. We compare our proposed method against a common black-box identification method using a deep neural network and show that incorporating physical consistency significantly improves the sample efficiency and generalizability of the model. Finally, we validate our method on the Spot quadruped robot across various locomotion tasks, showcasing its accuracy and generalizability in real-world scenarios over different gaits.
\end{abstract}

\section{Introduction}
Accurate physics models are crucial for the successful application of modern optimal control \cite{wensing2023optimization} and reinforcement learning \cite{ha2024learning} algorithms on legged robots. While these models are typically derived from CAD models, discrepancies frequently arise due to manufacturing variances, assembly imperfections, and the integration of additional components, such as wiring and sensors, that are not reflected in the original design. Furthermore, certain aspects such as flexibility, joint friction, etc., are not captured in CAD models. 

While the problem of rigid-body parameter identification for manipulators is well understood \cite{atkenson1986,khalil2004modeling}, the problem is much more complicated for legged robots as these systems are under-actuated (as opposed to fixed-base manipulators) and engage in intermittent, uncertain contact interactions with the environment (as opposed to wheeled and aerial robots). Furthermore, ensuring the physical consistency of the identified parameters, residing on complex manifolds, adds further complexity.

Classically, system identification of legged robots have largely relied on full Force/Torque (F/T) sensing, utilizing either onboard F/T sensors at the end-effectors \cite{Yusuke2014} or data collected from F/T plates during locomotion \cite{Ayusawa2013}. However, these methods have practical limitations; not all legged robots are endowed with full F/T sensors, and F/T plates are often impractical for real-world applications. Indeed, most state-of-the-art quadrupedal \cite{katz2019mini,grimminger2020open, SpotWebsite} and bipedal \cite{chignoli2021humanoid,cassie,daneshmand2021variable} platforms can \emph{only} measure or estimate joint torques (from motor current), without F/T sensors at end-effectors. Hence, a new inertial parameter identification framework that relies only on joint torque data is desired.
\begin{figure}[!t]
	\centering
	\includegraphics[width=1.0\linewidth, trim=0 120 0 0, clip]{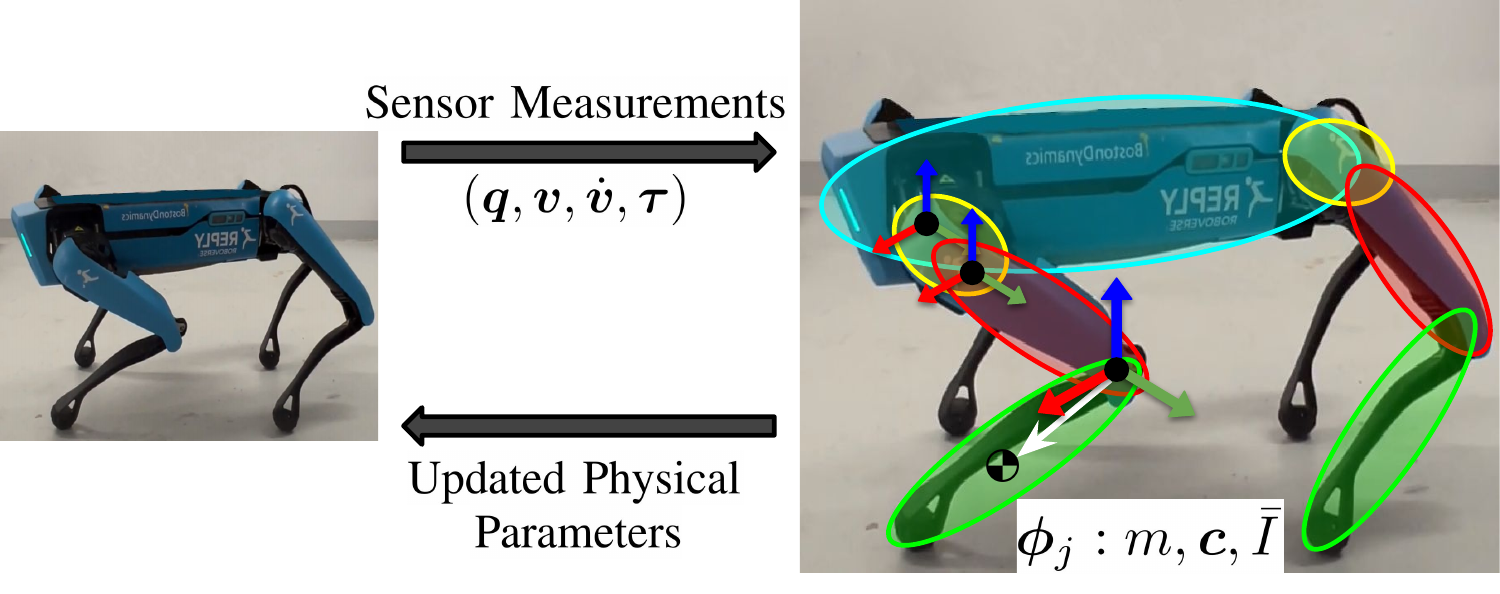}
        \vspace{-5mm}
	\caption{Whole-body inertial parameter identification —mass, center of mass and moment of inertia matrix— of floating-base robots using constraint-consistent dynamics with joint torque measurements and constrained convex optimization.}
	\vspace{-5mm}
	\label{fig:cover_image}
\end{figure}
Our previous work has successfully used a constraint-consistent dynamics model for centroidal state estimation of legged robots \cite{Khorshidi2023}. Using the same model, here we formulate the inertial parameter identification problem as a Linear Matrix Inequality (LMI), relying only on joint torque measurements. We leverage the projected dynamics to accurately identify inertial parameters —namely, mass, Center of Mass (CoM) position, and the moment of inertia matrix— of rigid body chains in legged robots~(Fig.~\ref{fig:cover_image}).
The main contributions of this work are as follows:
\begin{itemize}
    \item We introduce a novel approach for inertial parameter identification in legged robots by projecting the whole-body dynamics into the null space of contact constraints, enabling identification using only joint torque measurements. Our method extends the LMI formulation in \cite{Patrick2018lmi} to under-actuated systems with intermittent contact with the environment.
    \item Our extensive simulations on the Solo12 quadruped demonstrate that, unlike black-box identification, ensuring physical consistency significantly improves generalizability across new motions and tasks, without retraining and using more data.
    \item Through experiments on the Boston Dynamics Spot~\cite{SpotWebsite} across a variety of locomotion tasks, we showcase the accuracy and generalizability of our model in the real world. 
\end{itemize}
\section{Related Work}
Early research on inertial parameter identification focused on fixed-base robots, such as manipulators and industrial robots. Notable works include the foundational studies on inertial parameter identification by \cite{atkenson1986,khalil2004modeling}, which provided a basis for subsequent payload inertial parameter identification \cite{Wisama2007payload}. These efforts often emphasized designing trajectories that optimally excite the dynamics for accurate parameter identification \cite{Swevers1997excitation}.

With the development of more complex robotic systems, particularly legged robots with floating bases, research has shifted toward addressing the unique challenges posed by these platforms, such as contact switches, limited range of motion during identification, and under-actuation of the robot base. The identifiability of a minimal set of inertial parameters, also known as base inertial parameters, was demonstrated in \cite{Ayusawa2013} by leveraging only the unactuated dynamics of the floating base. Later works, like~\cite{Yusuke2014}, used the whole-body dynamics of the floating base for parameter identification, assuming the availability of full contact F/T sensing and joint torque measurements. In addressing the challenge of inertial parameter estimation with partial F/T sensing, \cite{Mistry2009partial} identified a subset of sensors and utilized Singular Value Decomposition (SVD) techniques to identify this subset for identifiability of base parameters. Unlike \cite{Yusuke2014} and \cite{Ayusawa2013}, which require F/T sensors for parameter identification, our approach relies solely on joint torque measurements, eliminating the need for contact force data. Although \cite{Ruben2018contactInvariant} also employs the same approach to remove the dependency on contact forces, their work focuses on learning residual model errors for improved inverse dynamics control rather than identifying inertial parameters. 
Moreover, by formulating the identification problem as a constrained optimization using LMI formalism, we ensure physical consistency—an aspect often neglected in traditional unconstrained least squares methods.



Constrained optimization has emerged as an important tool to improve the physical feasibility of the identified parameters. \cite{Jovana2016hierarchical} introduced a constrained hierarchical optimization technique that incorporates constraints such as positive mass, a convex hull for the CoM, and the positive definiteness of the moment of inertia matrix. One of the earliest attempts to use LMI formalism for ensuring the physical consistency of inertial parameters was demonstrated by \cite{Sousa2014lmi} on a seven-link robot. However, while this method could enforce positive definiteness of the moment of inertia matrix, it did not address more complex constraints, such as triangle inequalities, which arise from considering principal moments of inertia in rigid bodies.
Building on these efforts, \cite{Silvio2016manifold} proposed a manifold parameterization of inertial parameters to handle these constraints. However, the resulting optimization is nonlinear and lacks guaranteed global optimality. To address these limitations, \cite{Patrick2018lmi} advanced LMI formulations by using the pseudo inertia matrix to incorporate triangle inequalities, demonstrating promising results on the MIT Cheetah leg. However, \cite{Patrick2018lmi} treated the problem as a fixed-base identification for the leg. In this work, we extend the LMI formulation for whole-body inertial parameter identification of floating-base robots during locomotion tasks. 

More recently, researchers have recognized that inertial parameters reside on a curved Riemannian manifold \cite{Taeyoon2018geometric}, which complicates the optimization process. In their work, the authors have shown that regularization techniques based on coordinate-invariant distances improves accuracy and robustness in the identification process, however leading to non-convex and computationally expensive optimization problems. To overcome these computational challenges while maintaining robustness against noise, \cite{Taeyoon2020convex} proposed a second-order approximation of the geodesic distance as a regularization term, achieving convexity without sacrificing performance. We leverage this regularization term to improve the robustness of our identification while being computationally efficient.
\section{Fundamentals}
\subsection{Notation and Definitions}
\begin{itemize}
	\item Throughout the paper, scalars and scalar-valued functions are denoted by lowercase letters, vectors and vector-valued functions by bold lowercase letters, and matrices by uppercase letters.
	\item $\identity_n \in \mathbb{R}^{n \times n} $ is the identity matrix.
	\item $A^\dagger$ stands for the Moore-Penrose inverse of matrix $A$.
	\item $\text{tr}(A)$ denotes the trace operator over matrix $A$, which is the sum of its diagonal elements.
	\item $A\succcurlyeq 0$ and $A\succ0$ indicate that the symmetric matrix $A$ is positive semi-definite and positive definite, respectively.
	\item $\left\|\boldsymbol{x}\right\|_2^2$ denotes the squared Euclidean norm of vector $\boldsymbol{x}$.
	\item $\left\| \boldsymbol{x} \right\|_{Q}$ represents the weighted Euclidean norm of vector $\boldsymbol{x}$, such that $\left\| \boldsymbol{x} \right\|^2_{Q} = \boldsymbol{x}^\top Q\: \boldsymbol{x}$, where $Q \succ 0$.
\end{itemize}
\subsection{Floating-Base Dynamics}
The dynamics of a floating-base system with $n$ actuated joints can be expressed as
\begin{equation}\label{eq:rigid_body}
    M(\gencoor) \dot \genvel + \nonlinear(\gencoor,\genvel) = S^{\top} \torque + J_c^\top \force,
\end{equation}
where the variables are defined as follows:
\begin{itemize}
	\item $M(\gencoor) \in \mathbb{R}^{(n+6) \times (n+6)}$ is the mass-inertia matrix.
	\item $\gencoor \in \mathit{SE}(3) \times \mathbb{R}^n$ denotes the configuration vector.
	\item $\genvel \in \mathbb{R}^{n+6}$ encodes the vector of generalized velocities.
	\item $\nonlinear(\gencoor,\genvel) \in \mathbb{R}^{n+6}$ represents the nonlinear terms, which include centrifugal, Coriolis, and gravitational effects.
	\item $S=[0_{n\times6} \quad \identity_n] \in \mathbb{R}^{n \times (n+6)}$ is the selection matrix that separates the actuated and unactuated degrees of freedom.
	\item  $\torque \in \mathbb{R}^n$ is the vector of joint torques.
	\item $J_c \in \mathbb{R}^{3n_e \times (n+6)}$  is the Jacobian matrix corresponding to the $n_e$ contact points (e.g., feet). 
	\item $\force \in \mathbb{R}^{3n_e}$ is the vector of contact forces.
\end{itemize}
Although the formulation assumes point-contact feet for quadrupeds, it is equally applicable to humanoids with flat feet by extending it to account for the full contact wrenches.
\subsection{Motion and Constraint Dynamics}
Assuming rigid, non-slipping contact with the environment, we use the orthogonal projection operator $\small{\nullspace = \identity_{m} - J_c^{\dagger} J_c}$, with $m=n+6$, to project the dynamics into the null space of the contact Jacobian. This allows us to decompose the dynamics described in \eqref{eq:rigid_body} into the following set of equations \cite{mistry2012operational}
\begin{subequations}
\label{eq:rigid_body_separation}
\begin{align}
    \nullspace (M(\gencoor) \dot \genvel + \nonlinear(\gencoor,\genvel)) &= \nullspace S^\top \torque, \label{subeq:motion_space}\\
    \!\!(\identity_{m} \! -\! \nullspace)(M(\gencoor) \dot \genvel + \nonlinear(\gencoor,\genvel)) &= (\identity_{m} \! -\! \nullspace)S^\top \! \torque + J_c^\top \! \force, \label{subeq:constraint_space}
\end{align}
\end{subequations}
where \eqref{subeq:motion_space} encodes the dynamics of the robot's motion independently of the constraint forces, whereas \eqref{subeq:constraint_space} represents the dynamics within the contact constraint space~\cite{mistry2012operational}. In other words, $\nullspace S^\top \torque$ governs the system’s movement along a desired trajectory without violating the contact constraints, and $(\identity_m - \nullspace) S^\top \torque$ preserves the stationary contact. Notably, since \eqref{subeq:motion_space} is independent of contact forces,  we can use it to relate the measured joint torques to the joint accelerations without knowledge of the contact forces. This is particularly advantageous for robots without F/T sensors at end-effectors. In this work, we leverage these dynamics for the purpose of inertial parameter identification through contact.
\section{Inertial Parameter Identification}
\subsection{Problem Statement}
We aim to identify the rigid body inertial parameters of a floating-base robot. For a single rigid body, these parameters are defined as
\begin{equation}\label{eq:inertial_vector}
	\boldsymbol{\phi}_j = [m, h_x, h_y, h_z, i_{xx}, i_{xy}, i_{xz}, i_{yy}, i_{yz}, i_{zz}]^\top \in\mathbb{R}^{10},
\end{equation}
where $m$ represents the body mass, $\boldsymbol{h}=[h_x, h_y, h_z]^\top = m\com$ represents the first moment of inertia, $\com \in \mathbb{R}^3$ being the CoM position in the body frame, and 
\begin{equation}
	\bar{I} = \begin{bmatrix}
		i_{xx} &i_{xy} &i_{xz}\\
		i_{xy} &i_{yy} &i_{yz}\\
		i_{xz} &i_{yz} &i_{zz}\\
	\end{bmatrix} \in \mathbb{R}^{3\times3}, \quad \bar{I} \succ 0,
\end{equation}
is the moment of inertia matrix expressed in the body frame about the joint origin.

As shown in \cite{atkenson1986}, the complete dynamics of a robotic system, given by \eqref{eq:rigid_body}, can be expressed linearly w.r.t. the set of inertial parameters $\inertial$ as follows
\begin{equation}\label{eq:llsq}
	Y(\gencoor, \genvel, \dot \genvel) \inertial = S^{\top} \torque + J_c^\top \force,
\end{equation}
where $\inertial = [\inertial_1^\top, \inertial_2^\top,\: \dots \:, \inertial_{n_b}^\top]^\top \in \mathbb{R}^{10n_b}$ is the vector of inertial parameters for the $n_b$ links of the floating-base robot, and $Y(\gencoor, \genvel, \dot \genvel)$ is the regressor matrix computed from the kinematic information. By projecting these dynamics into the null space of contact points (as in \eqref{subeq:motion_space}), we can rewrite~\eqref{eq:llsq} as
\begin{equation}\label{eq:proj_llsq}
	P Y(\gencoor, \genvel, \dot \genvel) \inertial = \nullspace S^\top \torque.
\end{equation}

This equation leads to a linear least-squares minimization problem w.r.t. $\inertial$. In the following, we formulate this problem as an LMI~\cite{Patrick2018lmi} to ensure the physical feasibility of the identified parameters and robustness against noisy measurements.
\subsection{Physical Consistency}
To guarantee that the identified inertial parameters are physically consistent, we impose a set of constraints on the identification problem, formulated as LMIs. These constraints focus on key physical properties of rigid bodies such as the CoM position, the structure of the pseudo-inertia matrix, and the realizability of mass distribution within a bounding ellipsoid.
\subsubsection{First Moment of Inertia Constraint}
We assume that a rigid body is confined within a given ellipsoid $E \subset \mathbb{R}^3$, described as \cite{boyd1994linear}
\begin{equation}\label{eq:ellipsoid}
	E = 
	\left\{
	\boldsymbol{x} \in \mathbb{R}^3\mid(\boldsymbol{x}-\boldsymbol{x}_c)^\top Q_s^{-1} (\boldsymbol{x}-\boldsymbol{x}_c) \leq 1
	\right\},
\end{equation}
where $\boldsymbol{x}_c$ represents the ellipsoid center, and $Q_s\!\in\!\mathbb{R}^{3\times3}$, $Q_s\succ0$ represents its shape as a diagonal matrix whose diagonal elements are the squares of the ellipsoid semi-axes $\boldsymbol{x}_s$. To ensure that the CoM of the rigid body lies within this ellipsoid, we can rewrite \eqref{eq:ellipsoid} for the CoM position as follows
\begin{equation}\label{eq:com_ellipsoid}
    m \left(1 - (\com - \boldsymbol{x}_c)^\top Q_s^{-1} (\com - \boldsymbol{x}_c)\right) \geq 0.
\end{equation}

Since $m Q_s \succ 0$, we impose \eqref{eq:com_ellipsoid} as an LMI constraint on the inertial parameters $\inertial$ by using the Schur complement condition for positive semi-definite matrices~\cite{boyd1994linear}
\begin{equation}
	C(\inertial_j)=
	\begin{bmatrix}
		m &(\boldsymbol{h}-m\boldsymbol{x}_c)^\top\\
		\boldsymbol{h}-m\boldsymbol{x}_c &mQ_s
	\end{bmatrix} \succcurlyeq 0 .
\end{equation}
\subsubsection{Second Moment of Inertia Constraints} In order to address complex constraints on the second moment of inertia, we introduce the pseudo inertia-matrix defined as~\cite{Atchonouglo2008}
\begin{equation}
	J(\inertial_j) = \begin{bmatrix}
		K & \boldsymbol{h} \\
		\boldsymbol{h}^ \top & m
	\end{bmatrix}  \in \mathbb{R}^{4\times4},
\end{equation}
where $K = \frac{1}{2}\text{tr}(\bar{I})\identity_3 - \bar{I}$, and there exists a 1-1  linear correspondence between $K$ and $\inertial$ such that, $\small{\bar{I} = \text{tr}(K)\identity_3 - K}$~\cite{Patrick2018lmi}.

Additionally, since \eqref{eq:ellipsoid} is polynomial w.r.t. $\boldsymbol{x}$, we can express this ellipsoid as follows
\begin{equation}
	E = 
	\left\{
		\boldsymbol{x} \in \mathbb{R}^3 \mid
		\left[\begin{smallmatrix} \boldsymbol{x} \\ 1 \end{smallmatrix}\right]^\top Q_j \left[\begin{smallmatrix} \boldsymbol{x} \\ 1 \end{smallmatrix}\right] \geq 1 
	\right\},
\end{equation}
where
\begin{equation}
    Q_j = 
    \begin{bmatrix}
        -Q_s^{-1} & Q_s^{-1}\boldsymbol{x}_c\\
        (Q_s^{-1}\boldsymbol{x}_c)^\top & 1-\boldsymbol{x}_c^\top Q_s^{-1} \boldsymbol{x}_c
    \end{bmatrix} \in \mathbb{R}^{4\times4}.
\end{equation}

Finally, for a rigid body to be considered physical fully-consistent, two key conditions must be met: positive definiteness of the moment of inertia matrix while satisfying the triangle inequalities, and realizable mass distribution within the bounding ellipsoid $E$, preventing the rigid body from degenerating into a point mass. These conditions are satisfied if and only if the following constraints hold \cite{Patrick2018lmi}
\begin{equation}
	J(\inertial_j) \succ 0 \ \ \text{and} \ \ \text{tr}\left(J(\inertial_j) \; Q_j \right) \geq 0 .
\end{equation}
\subsection{Linear Matrix Inequalities Formulation}
Given $n_s$ data samples, we aim to optimize the following objective function, subject to the constraints, ensuring physical fully-consistent identification of inertial parameters~\cite{Taeyoon2020convex}
\begin{equation}\label{eq:lmi}
	\begin{aligned}
		\underset{
			\substack{ \inertial\\ B_v, B_c }
		}
		{\mathrm{min}}
		&\frac{1}{n_s}\! \sum_{k=1}^{n_s}\! \left\| \nullspace_k Y_k \boldsymbol{\phi}-\!\nullspace_k S^\top \! \left( \torque_k - B_v \mathbf{v}_k-\!B_c \: \text{sign}(\mathbf{v}_k)\right)\right\|_2^2 \\
		&+ \gamma \: || \boldsymbol{\phi} - \boldsymbol{\hat{\phi}} ||^2_{G} , \\
		\text{s.t.} \quad
		&C(\boldsymbol{\phi}_j) \succcurlyeq 0, \\ &J(\boldsymbol{\phi}_j) \succ 0, \\ &\text{tr}(J(\boldsymbol{\phi}_j) Q_j) \geq 0, \quad \forall j \in \begin{Bmatrix}
			1, \dots, n_b
		\end{Bmatrix}.
	\end{aligned}
\end{equation}

In this equation, $\mathbf{v}_k \in \mathbb{R}^n$ represents the joint velocities, while $B_v \in \mathbb{R}^{n\times n}$ and $B_c \in \mathbb{R}^{n\times n}$ are diagonal matrices representing viscous and Coulomb friction coefficients for the actuated joints, respectively. This formulation accounts for transmission losses by incorporating the friction coefficients into the projected least squares problem described by~\eqref{eq:proj_llsq}.

Additionally, to account for noisy data, a regularization term is added to the objective function. This term penalizes deviations from prior estimates of the inertial parameters $\small{\boldsymbol{\hat{\phi}} \in \mathbb{R}^{10n_b}}$, which are typically obtained from a CAD model of the robot. The regularization is weighted by matrix $G$, which approximates the geodesic distance in the space of inertial parameters. Since the mass-inertia parameters reside on a curved Riemannian manifold, using the geodesic distance provides a more accurate measure than a simple Euclidean distance. We adopt the second-order approximation of the geodesic distance, as introduced in \cite{Taeyoon2020convex}, ensuring the convexity of the optimization problem while accurately capturing the curvature of the parameter space.
\section{Experimental Results}
In this section, we evaluate the effectiveness of the proposed inertial parameter identification approach on two quadruped robots: Solo12 \cite{grimminger2020open} in a simulated environment with realistic added noise and Boston dynamics Spot \cite{SpotWebsite} in real-world scenarios. We also compare the performance of our proposed \emph{gray-box} constrained optimization identification against a \emph{black-box} neural network-based identification. These black-box models are frequently employed in model-based reinforcement learning for tasks such as locomotion~\cite{Nagabandi2017NeuralND} and manipulation \cite{Nagabandi2019DeepDM}. Although these models show promising results in predicting dynamic behaviors, they often exhibit limited generalizability across new tasks due to their \emph{black-box} nature that does not use the knowledge about the physics of the world (inductive bias). In contrast, our method integrates physical constraints, ensuring that the identified inertial parameters are physically consistent, with greater potential for generalization to out-of-distribution data and new tasks.
\subsection{Simulation}
\begin{figure}[t]
	\center
	\scalebox{1}{\input{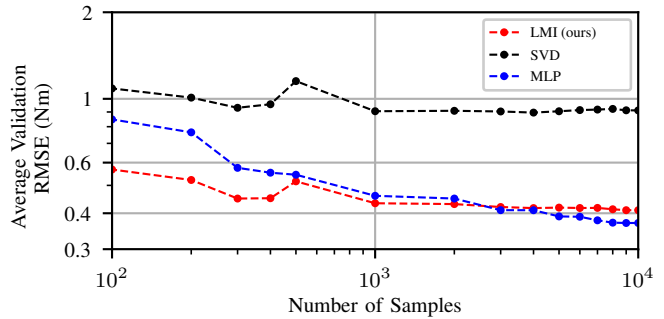}}
	\vspace{-4mm}
	\caption{Average RMSE comparison of torque predictions for the LMI (ours), SVD, and MLP models trained on different dataset sizes for Solo12 trot motion. Our LMI model consistently outperforms the MLP on smaller datasets, making it well-suited for real-world applications with limited number of samples.}
	\label{fig:solo_rmse}
	\vspace{-3mm}
\end{figure}
\begin{table}[b]
    \vspace{-2mm}
    \centering
    \caption{RMSE values of predicted torques (Nm) with LMI, SVD, and MLP models across different motions. Our LMI model demonstrate better generalizability across new motions and tasks.}
    \begin{tabular}{|c|c|c|c|}
        \hline
        Motion                 & LMI (ours)  & SVD   & MLP   \\ \hline
        Validation Dataset              & 0.4019 & 0.9867 & \textbf{0.3707} \\ \hline
        Out of Distribution Trot        & \textbf{0.6470} & 1.7184 & 1.0742 \\ \hline
        Out of Distribution Task (Jump) & \textbf{0.7148} & 3.4063 & 2.3986 \\ \hline
    \end{tabular} \label{tab:torque_rmse}
    \vspace{-2mm}
\end{table}
\begin{figure*}[t]
	\center
        \vspace{2mm}
	\scalebox{1}{\input{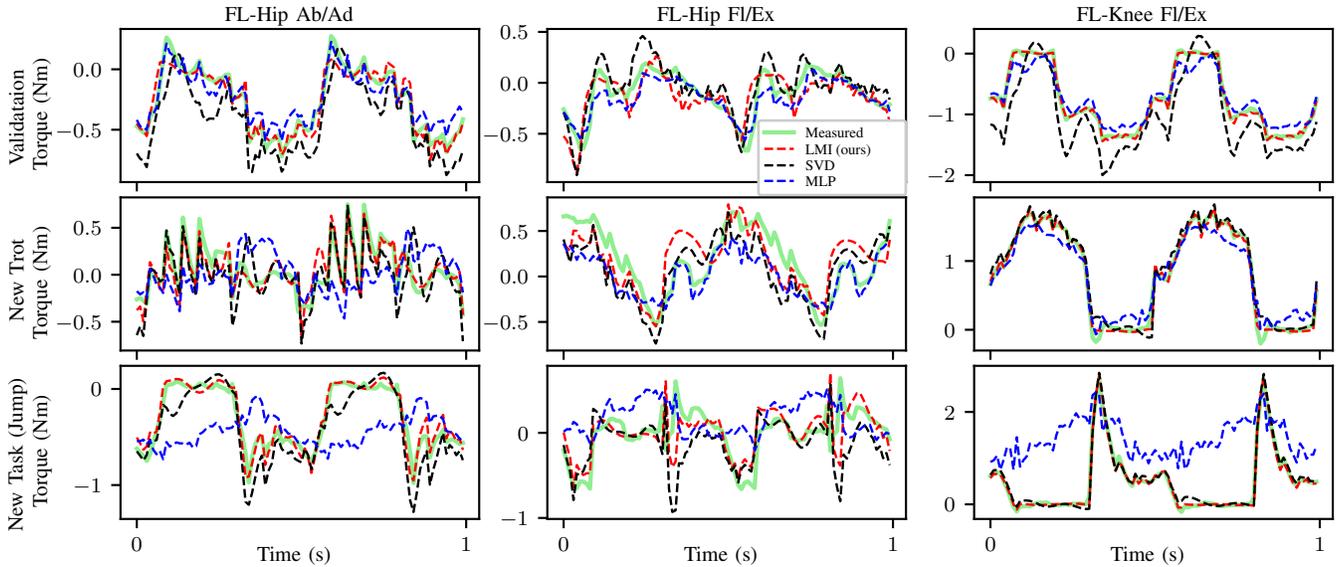}}
	\vspace{-4mm}
	\caption{Comparison of predicted and measured torques for the front left leg of Solo12 across three joint actuations. Our LMI model consistently delivers accurate torque predictions, even for the challenging jump motion. In contrast, while the Multi Layer Perceptron (MLP) model performs adequately on the validation dataset, it struggles to capture the complex dynamics of the jump, underscoring its limitations in generalizing to out-of-distribution tasks. For clarity, only $t=1$ s of the motion is displayed.}
	\label{fig:solo_fl_torque}
	\vspace{-6mm}
\end{figure*}
For this scenario, we use the Solo12 quadruped ($\small{m = 2.5}$~kg) in the PyBullet simulation environment \cite{pybullet2021}, executing a series of dynamic trot gaits generated by a whole-body motion planning framework \cite{avadesh2023mpc}. Data are sampled at 100 Hz, gathering a dataset of $n_s = 10^4$ samples that includes a variety of base velocities ($v_x, v_y \in [-1, 1]$ m/s) during the trot motion. We compare three models for identification problem:
\begin{itemize}
    \item LMI Model: Our proposed method, which solves the constrained LMI problem in \eqref{eq:lmi}.
    \item SVD Model: An unconstrained linear least-squares approach \cite{Ayusawa2013} based on SVD of \eqref{eq:proj_llsq}.
    \item MLP Model: A neural network model, in particular a multilayer perceptron, features a fully connected architecture with 5 hidden layers, each containing 2048 units, and employs \emph{LeakyReLU} and \emph{Tanh} activation functions to capture nonlinear dynamics. It follows a common end-to-end learning approach \cite{Nagabandi2019DeepDM} to predict actuation from sensor measurements.
\end{itemize}
\begin{figure*}[hb!]
	\centering
	\includegraphics[width=0.85\linewidth, trim=0 0 0 0, clip]{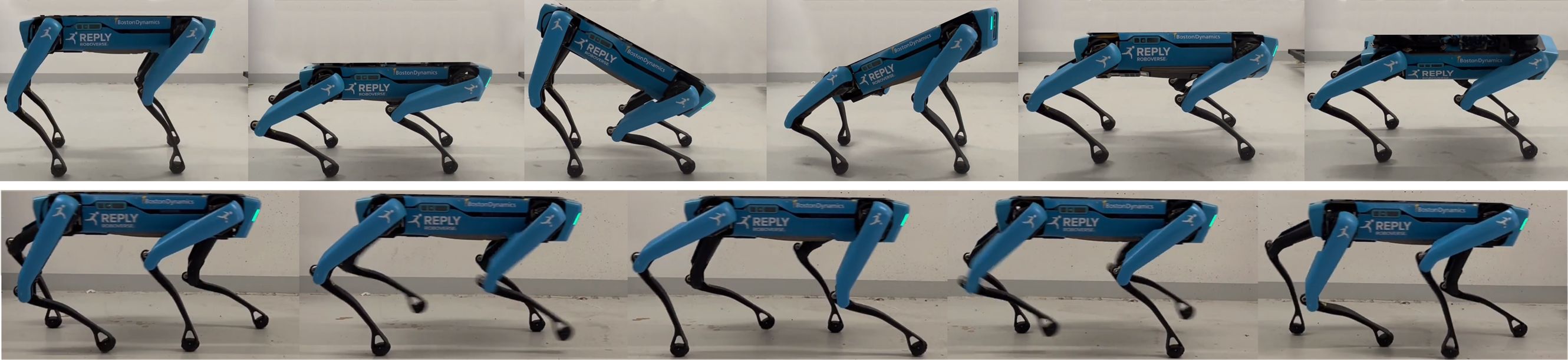}
        \vspace{0mm}
	\caption{Data collected from Spot across various locomotion tasks. Top: base motion with all feet in contact. Bottom: forward-backward walking.}
	\label{fig:spot_motion}
\end{figure*}

We train each model using different subsets of the dataset, ranging from $10^2$ to $10^4$ number of samples. The average Root Mean Squared Error (RMSE) for torque predictions across five validation datasets, each with a trajectory length of 30 seconds, is shown in Fig. \ref{fig:solo_rmse} on a log-log scale. The MLP model tends to perform better with large datasets, whereas our LMI model consistently outperforms the MLP model on smaller datasets, demonstrating its suitability for learning in the real world where limited data is available. The SVD model, which lacks constraints, exhibits higher prediction errors.

To evaluate generalizability without retraining, we test the fully trained models ($n_s = 10^4$) on out-of-distribution data, including:
\begin{itemize}
    \item A new trot motion with different base velocities, not included during training.
    \item A new locomotion task, jumping motion, distinct from any gait used during training.
\end{itemize}
The RMSE values for both the validation and out-of-distribution datasets are summarized in Table \ref{tab:torque_rmse}. Our LMI model maintains low prediction errors and demonstrates strong generalization to new motions. In contrast, the MLP model struggles with out-of-distribution generalization, especially for the jumping motion, leading to significantly higher errors. The SVD model, lacking physical consistency, shows higher prediction errors. Torque predictions for the front left leg of the three datasets—validation, out-of-distribution trot, and out-of-distribution jump—are depicted in Fig. \ref{fig:solo_fl_torque}. Our LMI model accurately predicts torques for all three joint actuations (hip abduction/adduction, hip flexion/extension, and knee flexion/extension), even for the more complex jumping motion. The MLP model, however, fails to capture the torque dynamics in this case.
\subsection{Real-World Experiment}
\begin{table*}[t]
	\centering
        \vspace{1mm}
	\caption{RMSE values of predicted torques (Nm) with LMI and MLP models across various motions on Spot quadruped. Our LMI model consistently outperforms the MLP model in both validation and new locomotion task. Torques are projected into the null space of the contact points. Joint order for each leg is hip abduction/adduction, hip flexion/extension, and knee flexion/extension. The robot’s total mass is $m = 34$ kg.}
	\begin{tabular}{|c|c|c|c|c|c|c|}
		\hline
		Motion & Model & Front Left & Front Right & Rear Left & Rear Right & Overall RMSE\\ \hline
		\multirow{2}{*}{\shortstack{Crawl \\ (Validation)}} & LMI (ours) & [1.015, 1.856, 3.089] & [1.037, 1.823, 3.180] & [1.452, 2.257, 3.506] & [1.113, 1.818, 3.928] & \textbf{8.246} \\
          & MLP & [1.981, 4.003,  3.558] & [1.256, 3.837,  4.681] & [2.197, 3.739, 5.489] & [2.093, 3.798, 5.606] & 13.039 \\ \hline\
		\multirow{2}{*}{\shortstack{Walk \\ (New Task)}} & LMI (ours)& [1.404, 2.549, 3.563] & [1.316, 2.829, 3.284] & [1.540, 2.585, 3.973] & [1.402, 2.450, 4.247] & \textbf{9.620}\\ 
        & MLP & [4.137, 4.443, 7.423] & [5.288,  3.343, 8.308] & [5.709, 3.597, 8.585] & [4.894, 4.592, 8.584] & 20.931\\ \hline
	\end{tabular}
	\label{tab:spot_rmse}
\end{table*}
\begin{figure*}[t]
	\center
	\scalebox{1.0}{\input{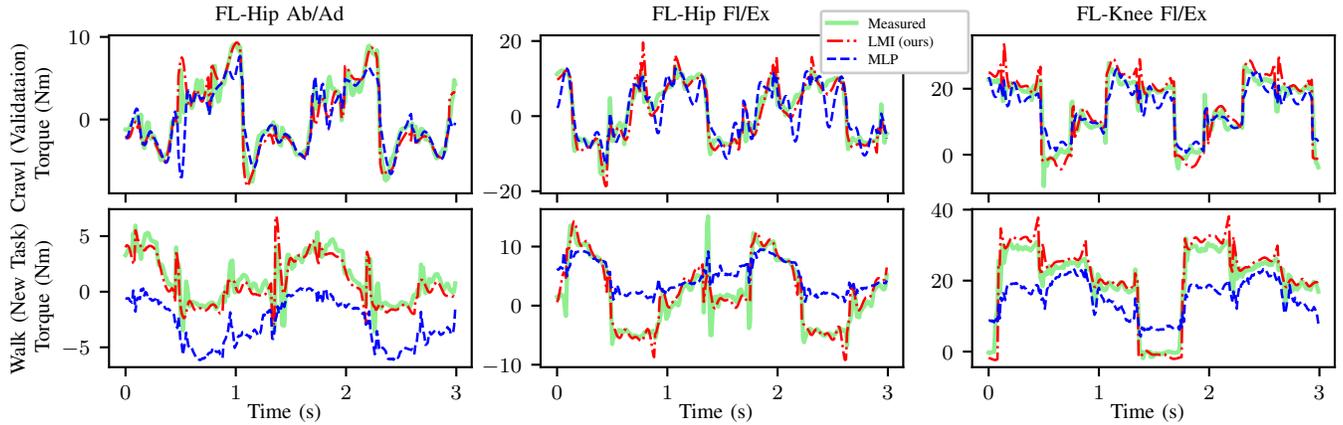}}
	\vspace{-4mm}
	\caption{Comparison of predicted and measured torques for the front left leg of Spot across three joint actuations. Our LMI model demonstrates accurate torque predictions for both validation and out-of-distribution tasks, while the MLP model struggles to generalize to new tasks. All torques are projected into the null space of the contact points.}
	\label{fig:spot_fl_torque}
	\vspace{-4mm}
\end{figure*}
We further applied our proposed identification method to the Spot quadruped robot \cite{SpotWebsite}, which is widely used for its reliability and adaptability across different domains. While the general specifications of Spot have recently become publicly available, custom configurations and modifications in various applications necessitate accurate model identification to develop tailored controllers. Notably, Spot does not come equipped with force/torque (F/T) sensors for contact force measurements, making it a relevant case study for system identification using only joint torque and kinematic data.

We interface with the robot using a control loop that publishes sensor measurements and state data at 100 Hz. Raw kinematic data, including generalized velocities, accelerations, and joint torque vectors, are post-processed using a forward-backward fifth-order low-pass Butterworth filter with a cutoff frequency of 10 Hz to mitigate noise. The dataset comprises various trajectories, such as base wobbling with all feet in contact, walking, and crawling at different base velocities and heights as shown in Fig.~\ref{fig:spot_motion}. A total of $n_s = 10^4$ samples from these three different motions are used for inertial parameter identification. Data from walking motions are reserved for validation to assess the model's generalization capabilities.

Initial estimates for the inertial parameters are based on the manufacturer's specifications \cite{bostondynamics} and adjusted for the total weight of our Spot robot, $m = 34$ kg. We compute the kinematic, contact Jacobian, and regressor matrices for each data sample using Pinocchio \cite{carpentier2019pinocchio}, and then solve the constrained linear least squares problem in \eqref{eq:lmi}, using MOSEK \cite{Erling2003} with a regularization factor of $\gamma = 10^{-2}$. Upon solving the identification problem, all physical consistency constraints were successfully satisfied. 

We further trained the MLP model using the same dataset and compared it with the LMI model in terms of torque prediction accuracy. The performance of both models was evaluated across two distinct motions, each with a trajectory duration of 30 seconds. Results for predicted torques are detailed in Table \ref{tab:spot_rmse} and visualized for the front left leg in Fig. \ref{fig:spot_fl_torque}, where all torques are projected into the null space of contact points in order to mitigate the influence of contact forces. The findings demonstrate the accuracy and generalizability of our LMI model in predicting joint torques, even when the robot operates under new motions not present in the training dataset. In contrast, the MLP model fails to capture the dynamics of the robot for new locomotion tasks.
\section{Conclusions and Future Work}
In this paper, we introduced a method for identifying rigid body inertial parameters in legged robots using joint torque measurements. By projecting the whole-body dynamics into the null space of the contact constraints, we eliminated the need for direct contact force/torque sensing. We then integrated the resulting dynamics into a convex optimization framework to ensure full physical consistency. We evaluated our identification approach against a widely used black-box MLP model, through simulations on the Solo12 quadruped with realistic noise conditions. Our results demonstrated that incorporating a physics-based model and enforcing physical consistency in the identification process significantly improves generalizability to new motions and tasks, offering clear advantages over black-box identification using neural networks. Furthermore, we validated our method on the Spot quadruped across various locomotion tasks, highlighting its accuracy and robustness in real-world scenarios. 

In the future, we aim to extend our framework to also include contact parameters in the identification problem. Integrating our parameter identification framework within an adaptive model predictive controller is also another avenue we aim to explore in our future work. Finally, we aim to explore loco-manipulation problems where the robot simultaneously adapts to the changes in its inertial parameters, contact properties, as well as object attributes.
\section{Acknowledgment}
Data collection, images, and videos in the real world experiment for this study were made possible with the use of a Spot robot provided by Roboverse Reply, Munich, Germany.
\bibliography{master} 
\bibliographystyle{ieeetr}

\end{document}